\documentclass[preprint,12pt]{elsarticle}

\usepackage{hyperref}
\usepackage{multirow}
\usepackage{amsmath}
\usepackage{algorithm}
\usepackage{longtable}
\usepackage{array}
\usepackage[noend]{algpseudocode}
\usepackage{natbib}
\usepackage[font=bf]{caption}
\usepackage{graphicx}
\usepackage{pgfplots}
\usepackage{xcolor}
\usepackage{subcaption}
\usepackage{amssymb}
\usepackage{setspace}
\usepackage[normalem]{ulem}
\useunder{\uline}{\ul}{}
\pgfplotsset{width=1.05\linewidth}
\algnewcommand{\algorithmicforeach}{\textbf{for each}}
\algdef{SE}[FOR]{ForEach}{EndForEach}[1]
  {\algorithmicforeach\ #1\ \algorithmicdo}% \ForEach{#1}
  {\algorithmicend\ \algorithmicforeach}% \EndForEach
\makeatletter
\def\BState{\State\hskip-\ALG@thistlm}
\makeatother
\journal{Information Sciences}
\begin{document}

\begin{frontmatter}

\title{Discrimination and Class Imbalance Aware Online Naïve Bayes}

%FUA-FL: Fairness-aware and Utility Agnostic for non-IID data in Federated Learning system

%FUA-FL: Fairness-aware and Utility Agnostic Federated Learning framework

%FUA-FL: Fairness-aware and Utility Agnostic Federated Learning framework for non-IID data


\author[mymainaddress]{Maryam Badar\corref{mycorrespondingauthor}}
\cortext[mycorrespondingauthor]{Corresponding author}
\ead{badar@l3s.de}

%\fntext[myfootnote]{Since 2021.}

\author[mymainaddress]{Marco Fisichella}
\ead{mfisichella@l3s.de}

\author[mymainaddress]{Vasileios Iosifidis}
\ead{iosifidis@l3s.de}

\author[mymainaddress]{Wolfgang Nejdl}
\ead{nejdl@l3s.de}
\address[mymainaddress]{L3S Research Center, Leibniz University of Hannover, Welfengarten 1, Hannover, 30167, Niedersachsen, Germany}
%Federated learning (FL) is an emerging communication-efficient system that is widely employed to address the privacy issues involved in distributed training. 
\begin{abstract}
Fairness-aware mining of massive data streams is a growing and challenging concern in the contemporary domain of machine learning. Many stream learning algorithms are used to replace humans at critical decision-making points e.g., hiring staff, assessing credit risk, etc. This calls for handling massive incoming information with minimum response delay while ensuring fair and high quality decisions. Recent discrimination-aware learning methods are optimized based on overall accuracy. However, the overall accuracy is biased in favor of the majority class; therefore, state-of-the-art methods mainly diminish discrimination by partially or completely ignoring the minority class. In this context, we propose a novel adaptation of Naïve Bayes to mitigate discrimination embedded in the streams while maintaining high predictive performance for both the majority and minority classes. Our proposed algorithm is simple, fast, and attains multi-objective optimization goals. To handle class imbalance and concept drifts, a dynamic instance weighting module is proposed, which gives more importance to recent instances and less importance to obsolete instances based on their membership in minority or majority class. We conducted experiments on a range of streaming and static datasets and deduced that our proposed methodology outperforms existing state-of-the-art fairness-aware methods in terms of both discrimination score and balanced accuracy.
\end{abstract}
%our method maintains parity between sensitive and non-sensitive groups for both majority and minority classes.
\begin{keyword}
\texttt online learning \sep discrimination-aware classification \sep class-imbalance aware classification
\end{keyword}

\end{frontmatter}

%\linenumbers
\section{Introduction}\label{sec1}
Enormous collections of continuously arriving data require efficient mining algorithms to render fair and  high quality predictions with minimum response delay. %Machine learning algorithms have far-reaching repercussions on global productivity, equity, inclusion, and environmental impact. Emerging machine learning based digital technologies have the potential to either hinder or enable global sustainable development. For instance, 
Many automated online decision-making systems have been proposed to supplement humans in several critical application areas subject to moral equivalence, such as credit risk assessment, online advertising, recruitment, and criminal recidivism assessment. These models have shown equivalently and in some cases better performance than humans. This argues for replacing human decisions with such models. However, such replacement has raised many challenging concerns regarding the fairness, transparency, and accountability of automated decision-making models.  Moreover, it has been empirically proven that intrinsically biased historical data has overtaken the digital world as well. In 2014, for example, Amazon developed an automated system to review job applicants' resumes to hire employees. In 2015, this automated reviewer was found to exhibit gender bias against women for technical positions \citep{dastinamazon}. Furthermore, the online travel company Orbitz was found to steer Mac users more to expensive hotels as compared to PC users \footnote{\url{https://edition.cnn.com/2012/06/26/tech/web/orbitz-mac-users/index.html}}. Machine learning technologies, applied in the field of medicine, are also found subject to biases hidden in the data \citep{pot2021not}. This has raised concerns regarding biases in the outcomes of such systems and their potentially harmful consequences. For example, a commercial health risk measurement algorithm was found exhibiting racial bias in its predictions \citep{obermeyer2019dissecting}. This system used "health care costs" as a proxy for "health care needs". Black patients, classified in the low risk category, had similar health needs as white patients who were classified in high risk category. Less money was spent on high-risk black patients than on high-risk white patients, resulting in incorrect learning of the algorithm.

Recent years have witnessed a plethora of studies that focus on detecting and mitigating discrimination embedded in the historical data. These learning algorithms ameliorate discrimination in a static fashion under the assumption that the characteristics of the data are not evolving. However, many real-world applications e.g., fraud detection, e-commerce websites, and stock market platforms, rely on real-time data streams. The real-time data evolves in a streaming fashion, and the statistical dependencies within the data also change over time (concept drift) \citep{liu2017regional}. The discriminatory outcomes have critical effects on current as well as future scenarios. For example, \citep{lelie2012european} suggests that even the second-generation of immigrants in Europe face ethnic disadvantages in employment compared to equally qualified Europeans. Thus, we need to detect and offset discrimination cumulatively while considering non-stationary nature of the data. Moreover, most fairness-aware learning approaches do not consider the inherent problem of class imbalance while mitigating bias. Class imbalance is the unequal distribution of instances across classes, i.e., the significantly skewed distribution of data towards one class (majority class). If the learning algorithm ignores the class imbalance problem when dealing with bias, there is a great possibility that the learner will achieve a low discrimination score by simply misclassifying the minority class instances. In such a case, the learner's low discrimination score is just a number that has nothing to do with its actual discrimination diminishing capability \citep{iosifidis2019adafair, iosifidis2020mathsf}. 

In this work, we propose a novel adaptation of Naïve Bayes that deals with class imbalance and diminishes discrimination simultaneously in a non-stationary environment. We have employed an Online Class imbalance Monitor (OCIM), which tracks the proportions of the majority and minority classes over the data stream. This monitor adjusts our proposed classifier to handle class imbalance. It has a smooth forgetting phenomenon that helps in handling concept drifts. We have also proposed a unique discrimination mitigation strategy that detects and offsets discrimination in a streaming environment. We have empirically proven that our class imbalance-aware and discrimination-aware classifier outperforms state-of-the-art fairness-aware learning methods. Moreover, we have also provided a detailed analysis of our results to elaborate the real cognizance behind the results.
\\
\\
\noindent Our key contributions are: 
\begin{itemize}
	\item A novel adaptation of Naïve Bayes to dynamically handle class imbalance (role of majority and minority class may swap over the stream) and mitigate discrimination as well as reverse discrimination (discrimination against the privileged group) over the stream.
	\item An extensive experimental evaluation of proposed model on a range of static and streaming datasets to prove the proposed method's superior predictive performance and discrimination mitigation capability.
\end{itemize}

\section{Related Work}
Our work encompasses four research areas: fairness-aware static learning, stream classification, fairness-aware stream learning, and class imbalance-aware learning.

\subsection{Fairness Aware Static Learning}
Literature provides many approaches for detecting and then diminishing discrimination. We can divide the discrimination mitigation strategies into three basic categories: Pre-processing, In-processing, and Post-processing techniques. This division depends on whether they modify the input training data, adapt the algorithm itself, or manipulate outputs of the model to mitigate discrimination.
 
\subsubsection{Pre-processing techniques:} 
The origins of the data have a decisive influence on the outcomes of decision-making models. If the origin of the data is prejudiced, then the decision-making model trained with the biased data will also behave prejudicially. Massaging \citep{kamiran2009classifying} is one of the most basic pre-processing techniques presented in the literature. It involves modification of class labels through minimal intrusive repercussions on the accuracy of the model. Reweighting \citep{calders2009building} is another less intrusive pre-processing method presented in literature to reduce discrimination. This method reduces discrimination by removing the dependence of model predictions on sensitive attributes by assigning weights to samples in training data. Weights are set based on the difference between the observed and expected probability of a sample with respect to a particular sensitive attribute and class. If the observed probability of a sample is lower than the expected probability, the sample is re-weighted with a higher weight. Preferential sampling \citep{kamiran2012data} is a special form of reweighting. It re-samples borderline objects with higher probability to minimize the adverse effect on predictive accuracy. The authors used a ranker to identify borderline objects in the training data. Data augmentation \citep{iosifidis2018dealing} is another potential method to deal with fairness. In this method, the authors proposed to over-sample minority group based on the sensitive attribute using Synthetic Minority Oversampling Technique (SMOTE). \citep{ijcai2018-430} proved that even if the training data is purely unbiased, discrimination still can exist in the predictions because pre-processing techniques cannot handle the bias introduced by the algorithm itself.

\subsubsection{In-processing techniques:}  
These techniques modify the classifier itself to obtain bias-free predictions. \citep{kamiran2010discrimination} presented a method to incorporate the condition of nondiscrimination into the objective function of their base model i.e., decision tree. \citep{aghaei2019learning} proposed a mixed integer programming based framework to achieve fair decision trees both for classification as well as regression. Furthermore, \citep{zafar2019fairness} provided a flexible convex-concave constraint based framework for fair margin based logistic regression classifier. Another in-processing approach to achieve a fair neural network based classifier is proposed by \citep{padala2020fnnc}. In this framework, the convex surrogates of constraints are included in the loss function of the neural network classifier through Lagrangian multipliers to achieve fairness. The literature also provides adaptive reweighting schemes to achieve fairness. For example, Adafair \citep{iosifidis2019adafair} is an Adaboost-based fairness-aware classifier designed to update instance weights in each boosting round, while considering the cumulative notion of fairness based on all members of the current ensemble.
%or use 
\subsubsection{Post-processing techniques:} 

These techniques alter the decisions of the classifier itself to diminish bias. For example, the authors of \citep{kamiran2010discrimination} have proposed a method which relabels certain leaves of a decision tree model to reduce discrimination while maintaining high predictive performance. \citep{calders2010three} provided a method to alter the probabilities of a naïve bayes classifier to tackle discrimination. In \citep{hajian2015discrimination}, the authors removed discrimination by processing the fair patterns with k-anonymity. \citep{fish2016confidence} proposed a method to alter the decision boundaries of an adaboost classifier to achieve fairness. \citep{NGUYEN2021542} presented a relabeling method based on Gaussian process that achieves fairness while maintaining high predictive accuracy.

\subsection{Stream Classification} 
 %A number of efforts have been made to adapt offline supervised learning methods for stream learning \citep{aggarwal2006framework, bifet2013efficient, losing2018tackling}.  
The main challenge in stream learning is to account for concept drifts. The model should be able to efficiently adapt to the changing patterns of data in the stream. The literature provides many batch learning methods for stream learning. For example, \citep{masud2012facing} proposed a semi supervised clustering method. Similarly, \citep{bifet2013efficient} presented a probabilistic adaptive windowing method for stream classification. The authors claim that their method improves the traditional windowing method because it includes older samples along with the new ones to maintain information regarding the previous concept drifts.These traditional batch learning methods lack the ability to continuously update the model with the arrival of each new sample.

Online learning avoids the cost of data accumulation. Moreover, online learning algorithms have the ability to converge faster compared to batch learning algorithms. \citep{chen2012online} present an online boosting algorithm, i.e., OSBoost, for classification in non-stationary environments. This algorithm is an adaptation of the offline SmoothBoost. The method proposed by \citep{krawczyk2019adaptive} is an adaptive active learning ensemble model that selects the most competent classifier from the pool of classifiers for each classification query. Another stream learning method is presented by \citep{YU2022996}. This method is developed to deal with concept recurrence with clustering. Whenever a concept recurs, the most appropriate model is retrieved from the repository and used for further classification. \citep{NGUYEN20191}  is another lossless learning classifier based on online multivariate Gaussian distribution (OVIG). An online version of semi supervised Support Vector Machine (SVM) is proposed by \citep{LIU2018125} which classifies newly arriving data based on few labelled instances of the data. 

\subsection{Fairness-aware stream learning}
This type of learning techniques reduce discrimination in streaming environment. A chunk-based pre-processing technique(Massaging) is proposed by \citep{iosifidis2019fairness} to mitigate discrimination. In this technique, the discrimination in each chunk is removed and then it is fed to the online classifier. FAHT (Fairness Aware Hoeffding Tree) \citep{zhang2019faht} is another method, based on decision tree, which is proposed to handle fairness in data streams. In this method, the notion of fairness is included in the attribute selection criteria for splitting the decision tree. The underlying decision tree grows by utilizing both information gain and fairness gain. FABBOO \citep{iosifidis2020mathsf} provides a method to change the decision boundary of the decision trees to achieve fairness. But FABBOO has kept the role of sensitive group fixed over the stream. It lacks the ability to handle reverse discrimination, i.e., the model starts discriminating towards the privileged group.

\subsection{Class imbalance-aware stream learning}
Class imbalance is an inherent problem of model learning. If the learning algorithm does not tackle class imbalance appropriately, it mostly learns by simply ignoring the minority class instances \citep{THABTAH2020429}. \citep{wang2016online} presented a cost-sensitive online learning algorithm based on bagging/boosting techniques for imbalanced data streams. Class imbalance can also be handled by instance weighting as proposed by FABBOO \citep{iosifidis2020mathsf}. Data augmentation is another potential method for handling class imbalance. For example, \citep{bernardo2020c} proposed a batch learning method, i.e., CSMOTE, to re-sample the minority class in a defined window of instances based on the synthetic minority oversampling technique (SMOTE).

\section{Problem Formulation}
In this work we are dealing with streaming data, so we need to continuously update the model. At every timepoint ${t}$, the instance ${x_{t}}$ (without label) is presented to the model for prediction, and later the label $y_t$ of instance ${x_{t}}$ is revealed to the model for training. This type of evaluation is called prequential evaluation \citep{gama2010knowledge} or test-then-train evaluation. The state-of-the-art continual learning models focus only on optimizing the model's overall accuracy, which leads to biased decision-making models. In our proposed continual learning model, we employ multi-objective optimization to reduce the discrimination score while dealing with the class imbalance problem. The main objectives of the proposed model are (i) fairness-aware online learning and (ii) class imbalance-aware online learning.

The proposed model is designed for binary classification. We assume that the streaming data has only one sensitive attribute with binary values. For example, if gender is a sensitive attribute, it can have two values {male and female}. The streaming data segregates into two portions based on the values of the sensitive attribute. Some proportion of the data belongs to the sensitive group (${S^{-}}$) and the rest of the data belongs to the non-sensitive group (${S^{+}}$). Primarily, the model is designed to give fair predictions about every new instance irrespective of the value of the sensitive attribute.    

We assess fairness in terms of a discrimination measure. There are many definitions of fairness in the literature \cite{verma2018fairness}, but so far, no clear criteria have been presented for choosing a particular fairness definition. In this research, we use the notion of statistical parity \cite{verma2018fairness} to assess the discriminating behavior of the proposed model. Statistical parity measures the difference in the probability of a subject being assigned to the positive class $y=y^{+}$ by the classifier, regardless of its membership in the sensitive $S^{+}$ or non-sensitive group $S^{-}$ as shown in equation \eqref{eq1}. Statistical parity does not take into account the true label of the subject and thus may lead to reverse discrimination, i.e., the model begins to discriminate against the privileged group. In our proposed model, we also address this reverse discrimination problem; the detail can be found in section \ref{Discrimination Aware Online Learning}. 

\begin{equation}\label{eq1}
    St.\: Parity = P(y=y^{+} \mid x \: \epsilon\:  S^{+}) - P(y=y^{+} \mid x \: \epsilon \: S^{-})
\end{equation}

We consider that the streaming population is unbalanced, i.e., one class dominates the other. The state-of-the-art models usually ignore the minority class, which leads to a large number of false negatives for the minority class \citep{calders2009building, calmon2017optimized, fish2016confidence, hardt2016equality, kamiran2009classifying}. The characteristics of the data change with every point in time i.e., the roles of the minority and majority classes may swap over time. Therefore, we present a generalized approach to deal with the class imbalance problem. 
\begin{figure}
  \includegraphics[width=\linewidth]{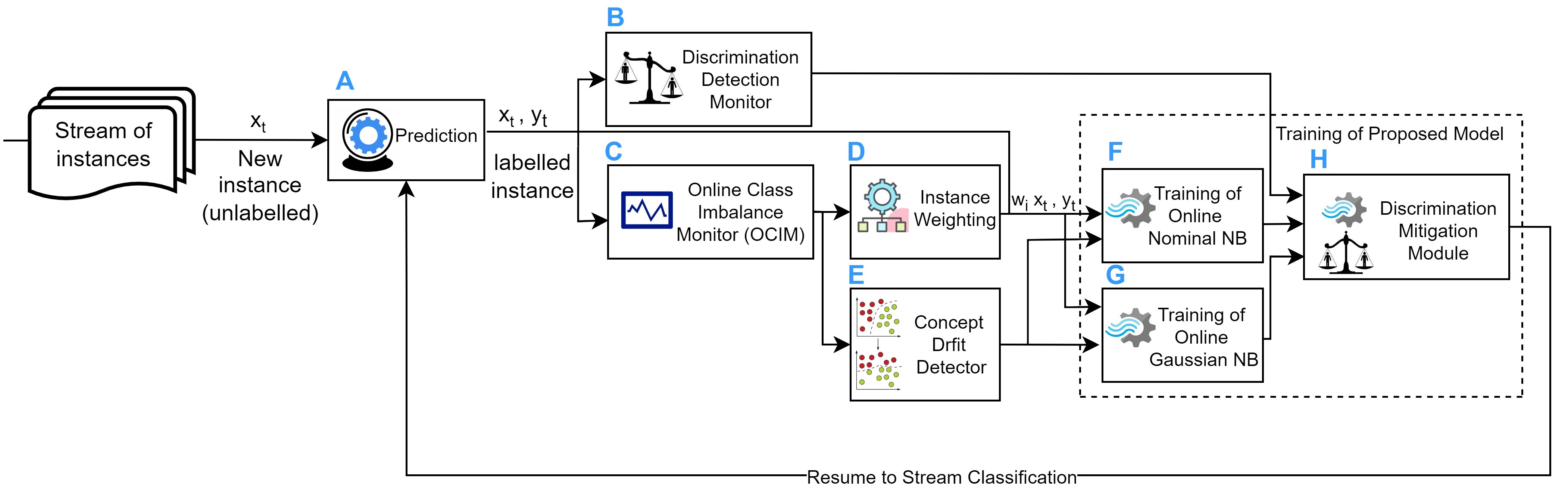}
  \caption{Illustration of proposed methodology}
  \label{fig:1}
\end{figure}
\section{Proposed Model}
An illustration of the proposed model is shown in Figure \ref{fig:1}. In this study, we are using prequential evaluation, therefore, as soon as a new instance $x_{t}$ arrives, it is tested using the proposed model (A). After testing, the instance $x_{t}$ with its true class label $y_{t}$ is fed to  discrimination detector (B) and online class imbalance monitor OCIM (C). The OCIM monitors the ratios of positive and negative classes throughout the stream and feeds the respective class ratios to instance weighting module (D). The instance weighting module adjusts instance weight in accordance with the respective class ratio to ensure class imbalance aware learning of proposed model. The class ratios obtained from OCIM are also used to keep track of the concept drifts using concept drift detector (E) and to handle concept recurrence. The instance $x_{t}$, its true label $y_{t}$ and the respective weight $w_{i}$ are used to train Online Nominal Naïve Bayes (F) and Online Gaussian Naïve Bayes (G). The discrimination detector monitors the discrimination over the stream using the notion of cumulative statistical parity and triggers the discrimination mitigation module (H) if cumulative statistical parity value exceeds a user defined threshold $\varepsilon$. Further details about these components are provided in the following subsections.

\subsection{Mixed Naïve Bayes}
In this work, we tailor the Naïve Bayes algorithm to process streaming data for which we do not have access to historical data. By default, Naïve Bayes is designed only for nominal data. However, in real life, datasets are usually a combination of nominal and continuous attributes. To accommodate both continuous and nominal attributes, we propose an online version of Mixed Naïve Bayes (MNB). MNB is a combination of online Nominal Naïve Bayes and online Gaussian Naïve Bayes. For each new instance, continuous attribute values are sent to online Gaussian Naïve Bayes and nominal attribute values are passed to online Nominal Naïve Bayes. Online Nominal Naïve Bayes and online Gaussian Naïve Bayes independently update themselves. The following sections illustrate the algorithmic details of these two models.

\subsubsection{Online Nominal Naïve Bayes}
 The proposed model is designed for binary classification only. Online Nominal Naïve Bayes maintains a summary for each class that contains the count of unique values of each nominal attribute. Whenever a new instance arrives, the summary is updated for the class to which the instance belongs. Since we are using prequential evaluation, thus, online Nominal Naïve Bayes model computes the posterior probabilities of each class with the arrival of each new instance using the equation \eqref{eq2} before updating the summaries.
 
\begin{equation}\label{eq2}
P(C \mid  a_{1},a_{2},a_{3},...,a_{n}) \sim P(C)\prod_{i=1}^{n}P(a_{i} \mid C) 
\end{equation}

\subsubsection{Online Gaussian Naïve Bayes}
Online Gaussian Naïve Bayes maintains the running mean and variance of each continuous attribute. For this purpose, we use Welford's online algorithm \citep{welford1962note}. %\footnote{https://github.com/isnarvaez/welford-online-algorithm}.
The running mean of each attribute is computed using the equation \eqref{eq3}. Here $\bar{a_{n}}$ is the current mean of the attribute, $n$ is the number of instances, $\bar{a}_{n-1}$ is the previous mean, and ${a_{n}}$ is the current value of the attribute. To calculate the variance, we need to calculate an intermediate term $M_{2,n}$ as shown in equation \eqref{eq4}. Once we have $M_{2,n}$, we can determine the running variance by equation \eqref{eq5}. With the arrival of every new instance in the stream, the online Gaussian Naïve Bayes updates each continuous attribute's running mean and variance. The summaries of continuous attributes contain the running mean and variance of the respective attribute. 

\begin{equation} \label{eq3}
\bar{a_{n}} =  \frac{(n-1)\bar{a}_{n-1}+a_{n}}{n}
\end{equation}

\begin{equation} \label{eq4}
M_{2,n} = M_{2,n-1} + (a_{n}-\bar{a}_{n-1})(a_{n}-\bar{a_{n}})
\end{equation}

\begin{equation} \label{eq5}
\sigma_{n}^{2} = \frac{M_{2,n}}{n}
\end{equation}
As we are using prequential evaluation, therefore, online Gaussian Naïve Bayes model computes the posterior probability of each class using the equation \eqref{eq5} before updating the running mean and variance. The only difference is in computing the likelihood of each attribute $a_{i}$, which is calculated using equation \eqref{eq6}.\\

\begin{equation}\label{eq6}
P(a_{i}\mid C) = \frac{1}{\sigma_{n} \sqrt{2\pi }}\: exp \left (  -\frac{(a_{n}- \bar{a_{n}})}{2\sigma_{n} ^{2}} \right )
\end{equation}

In the next sections, we describe the details of the modules we propose to handle class imbalance and discrimination in data streams.
\subsection{Module for monitoring and handling class imbalance}
We use a class imbalance monitoring component that tracks the percentages of classes in the stream. The roles of majority and minority classes may swap as the stream evolves, i.e., a class that is in the minority at the current time may turn out to be the majority at a later time. We track the state of disequilibrium using the Online Class Imbalance Monitor (OCIM) \citep{wang2013learning} as shown in equation \eqref{eq7}. In this equation, $CP_{t}^{+}$ is the percentage of positive class at time $t$ and $CP_{t}^{-}$ is the percentage of negative class at time $t$. After the arrival of each new record, OCIM updates the percentage $CP_{t}$ of the respective class using equation \eqref{eq8}.

\begin{equation} \label{eq7}
OCIM_{t} = CP_{t}^{+} - CP_{t}^{-}
\end{equation}

\begin{equation} \label{eq8}
CP_{t}^{y} = \alpha \cdot CP_{t-1}^{y} + (1-\alpha) \cdot \mathbb{I}[y,y_{t}]
\end{equation}

The state of imbalance needs to be changed based on the most recent examples from the stream, and the impact of previous examples needs to be reduced. Therefore, we include a temporal decay factor $(0<\alpha < 1)$ to capture the change in disequilibrium quickly. This decay factor limits the effect of the historical data; therefore, $CP_{t}$ is adjusted based on the most recent records. $\mathbb{I}[y,y_{t}]$ is the identity function that returns the value '1' if the predicted label ($y_{t}$) and the true label ($y$) are the same, otherwise it returns the value '0'.
\begin{algorithm}[t]
\begin{algorithmic}[1]
  \Require{true class labels y, positive class weight $CW^{+}$, negative class weight $CW^{-}$, $OCIM_{t}$ }
    \State{Initialize: current instance's weight $w_{i} = 1$;}
    \If{$y == negative\; label$ and $OCIM_{t} > 0$} 
        \State {$w_{i} = CW^{-} / (1-OCIM_{t})$}
    \EndIf
    \If{$y == positive \; label$ and $OCIM_{t} < 0$}
        \State {$w_{i} = CW^{+} / (1+OCIM_{t})$}
    \EndIf
\caption{Computing instance weights}
\label{alg:1}
\end{algorithmic}
\end{algorithm}
Once we have the class percentages (i.e., $CP_{t}^{+}$, $CP_{t}^{-}$), we can use them to find an appropriate weight for each new instance of the data stream. Algorithm~\ref{alg:1} presents the complete methodology for computing the instance weights. $CW^{+}$ and $CW^{-}$ are the class weights of the positive and the negative class, respectively. We compute $CW^{+}$ and $CW^{-}$ using the class weights library of Sklearn \footnote{\url {https://scikit-learn.org}, Accessed in 2022}.
%\citep{class_weights}.
%\footnote{https://scikit-learn.org, access in January 2022}.
%To handle class imbalance, we need to apply weighting scheme proposed in algorithm~\ref{alg:1} while updating the class summaries after the arrival of each new record.

\subsection{Module for handling concept recurrence}
 As shown in Figure \ref{fig:1}, we use a concept drift detector inspired by Page Hinkley's \citep{serakiotou1987change} explicit drift detection method. Our concept drift detection method monitors the OCIM parameter. This method of drift detection works by comparing the current OCIM to $OCIM\_mean_t$. $OCIM\_mean_t$ is the mean value of the OCIM computed for a window of instances up to the current time as illustrated in equation \eqref{eq9}. We chose a window of $1000$ instances to compute $OCIM\_mean_t$. In general, concept drift is detected when the observed $OCIM_t$ is above the mean $OCIM\_mean_t$ by a specified threshold $\eta$ at a given point in time. We chose the value of $\eta$ as $0.02$.
 
 \begin{equation} \label{eq9}
    OCIM\_mean_t = \frac{\sum_{i=1}^{N} OCIM_{_{i}}}{N}
\end{equation}

 Concept recurrence is a special case of concept drift where the concepts which are already seen in the past reappear in the evolving stream. As soon as concept drift is detected, the MNB stores the summaries of next instances as a separate model. In future, when similar concept reoccurs (similar concept drift recurs) then MNB retrieves the corresponding model and uses it for further prequential evaluation.
 
\subsection{Online Discrimination Detection and mitigation} \label{Discrimination Aware Online Learning}
We need to handle discrimination embedded in data streams. As the streams progress, the discriminated groups and the preferred groups do not remain the same. The group that was once discriminated against may turn out to be a preferred group later. Therefore, we need to develop a method that efficiently deals with this concept deviation. Also, we need to maintain the methodology that we developed to deal with the class imbalance problem. 

The discriminatory models have very long-lasting consequences, affecting not only current outcomes but also future outcomes. Short term discrimination detection methods fail to ameliorate discrimination over time because discrimination scores that are minor at a single time point may aggregate into considerable prejudice in the long run. Thus, in contrast to the short-term discriminatory measures applied by state-of-the-art stream learning methods, it is necessary to consider discriminatory outcomes cumulatively. We use the notion of cumulative statistical parity proposed by \citep{iosifidis2020mathsf} to detect and measure discrimination over the stream. Equation \eqref{eq10} illustrates the notion of cumulative statistical parity.

\begin{equation}\label{eq10} 
\frac{\sum_{i=1}^{t} \mathbb{I}[y_{t}=y^{+} \mid x_{i}\ \epsilon \ S^{-}]}{\sum_{i=1}^{t} \mathbb{I}[ x_{i}\ \epsilon\  S^{-}] + \gamma } - \frac{\sum_{i=1}^{t} \mathbb{I}[y_{t}=y^{+} \mid x_{i}\epsilon S^{+}]}{\sum_{i=1}^{t} \mathbb{I}[ x_{i} \ \epsilon \  S^{+}] + \gamma }
\end{equation}

The cumulative statistical parity is updated after the arrival of each new instance in the stream. '$\gamma$' is the adjustment factor, which is used to adjust the discrimination score at the beginning of the stream.
\begin{algorithm}[h]
\begin{algorithmic}[1]
  \Require{Summaries of the number of samples belonging to the positive class with sensitive value N($C_{+},S^{-}$); the number of samples belonging to the positive class with non-sensitive value N($C_{+},S^{+}$); the number of samples belonging to the negative class with sensitive value N($C_{-},S^{-}$); the number of samples belonging to the negative class with non-sensitive value N($C_{-},S^{+}$); discrimination score disc.}
  \Ensure{The number of samples does not change}
    \If{$disc > \varepsilon  $}
        \State $N(C_{+},S^{-}) = N(C_{+},S^{-}) + \lambda \: × N(C_{-},S^{-})$
        \State $N(C_{-},S^{-}) = N(C_{-},S^{-}) - \lambda \: × N(C_{-},S^{-})$
        \State $N(C_{-},S^{+}) = N(C_{-},S^{+}) + \lambda \: × N(C_{+},S^{+})$
        \State $N(C_{+},S^{+}) = N(C_{+},S^{+}) - \lambda \: × N(C_{+},S^{+})$
    \EndIf        
    %change indentation
    \If{$disc < -\varepsilon  $}
        \State $N(C_{+},S^{+}) = N(C_{+},S^{+}) + \lambda \: × N(C_{-},S^{+})$
        \State $N(C_{-},S^{+}) = N(C_{-},S^{+}) - \lambda \: × N(C_{-},S^{+})$
        \State $N(C_{-},S^{-}) = N(C_{-},S^{-}) + \lambda \: × N(C_{+},S^{-})$
        \State $N(C_{+},S^{-}) = N(C_{+},S^{-}) - \lambda \: × N(C_{+},S^{-})$
    \EndIf           
\caption{Online discrimination mitigation procedure}
\label{alg:2}
\end{algorithmic}
\end{algorithm}

To eliminate the discrimination, we change the probability distributions of the sensitive group $P(S^{-} \mid class)$ and the non-sensitive group $P(S^{+} \mid class)$ after the arrival of each new example in the data stream. If the discrimination value is greater than a certain threshold $\varepsilon$, we add a factor of the number of samples belonging to the negative class with sensitive value N($C_{-},S^{-}$) to the number of samples belonging to the positive class with sensitive value N($C_{+},S^{-}$). We also subtract the same factor from the number of samples belonging to the negative class with sensitive value N($C_{-},S^{-}$).% We only change the distributions of the samples with sensitive and non-sensitive values of the sensitive attribute. 
We do not want to increase the number of samples because we want to avoid unnecessary data augmentation.

Similarly, we add a factor of the number of samples belonging to the positive class with non-sensitive value N($C_{+},S^{+}$) to the number of samples belonging to the negative class with non-sensitive value N($C_{-},S^{+}$). We also subtract the same factor from the number of samples belonging to the negative class with the sensitive value N($C_{+},S^{+}$).

Since we want to deal with concept deviations in the evolving data streams, we also consider the negative discrimination, i.e., when the learner starts discriminating against the samples with non-sensitive value. To remove the negative discrimination, we use the same method as described above, except that now we swap the roles of sensitive and non-sensitive groups. Algorithm~\ref{alg:2} illustrates our online discrimination-aware learning procedure.
 
\section{Evaluation of Proposed Model}
We compare the proposed methodology against four baseline models including the class imbalance agnostic CSMOTE \citep{bernardo2020c}, non-stationary OSBoost \citep{chen2012online}, fairness agnostic Massaging (MS) \citep{iosifidis2019fairness}, fairness aware FAHT \citep{zhang2019faht}, and class imbalance and discrimination aware FABBOO \citep{iosifidis2020mathsf}. All baseline are trained using the same hyper-parameters as given in the respective research articles. We also evaluate different variants of MNB to stress on the importance of different modules of proposed model.

\begin{enumerate}
\item \textbf{CSMOTE} \citep{bernardo2020c}:  This baseline is not fairness agnostic, but it is designed to handle class imbalance in a non-stationary environment by re-sampling the minority class in a defined window of instances.
\item \textbf{OSBoost} \citep{chen2012online}: It is a classification model for data streams. It is not capable of handling either class imbalance or discrimination.
\item \textbf{Massaging (MS)} \citep{iosifidis2019fairness}: This is a fairness-aware learning method. It is a chunk based technique which handles discrimination in the current chunk by swapping labels. But it does not account for cumulative effects of discrimination, it is designed to handle discrimination only on short term basis i.e. for the current chunk. But it considers fairness based on its recent outcomes. We use their default chunk size for training this baseline i.e. $1000$. This method cannot handle class imbalance.
\item \textbf{Fairness Aware Hoeffding Tree (FAHT)} \citep{zhang2019faht}: This method is an adaptation of Hoeffding tree that is designed to deal with fairness. It incorporates the fairness gain along with the information gain into the partitioning criteria of the decision tree. This model is not able to deal with class imbalance and concept drifts.
\item \textbf{FABBOO} \citep{iosifidis2020mathsf}: This is an online boosting approach that handles class imbalance by monitoring class ratios in an online fashion. It employs boundary adjustment methods to handle discrimination.
\item \textbf{MNB (Mixed Naïve Bayes)}: It is a combination of nominal Naïve Bayes and Gaussian Naïve Bayes online learning model. It considers no notion of fairness and class imbalance while performing classification task.
\item \textbf{DAMNB (Discrimination Aware Mixed Naïve Bayes)}: This variant of MNB tackles both positive and negative (reverse) discrimination over the stream. 
\item \textbf{DCAMNB (Discrimination and Class imbalance Aware Mixed Naïve Bayes)}:  It is a variant of MNB which mitigates discrimination as well as handles class imbalance in the evolving stream. 
\end{enumerate}

\subsection{Datasets}
The details of the datasets used to test the efficiency of proposed model are shown in Table \ref{table1}. The datasets have different characteristics related to the number of attributes (\#Att.), number of instances (\#Inst.), sensitive attribute (Sens. Att.), and class ratio (positive to negative). We are using static datasets along with the streaming datasets. Despite the growing interest in AI models that focus on fairness, there is still a lack of large streaming datasets in this domain. Therefore, we use static datasets along with streaming datasets to prove the effectiveness of our proposed model. Since we are unaware of the temporal characteristics of the static datasets, we report the evaluation metrics on the average of $10$ random shuffles of each static dataset that passes through the model.
\begin{table}[t!]
\small
\caption{Description of Datasets}
\begin{center}
\begin{tabular}{p{2.2cm}p{1.2cm}p{1cm}p{1.8cm}p{1.6cm}p{1.8cm}p{1cm}}
\hline
Dataset & \#Inst. & \#Att. & Sens. Att. & \begin{tabular}[c]{@{}l@{}}Class Ratio \\ (+:-) \end{tabular} & Positive Class & Type \\
\hline
\begin{tabular}[c]{@{}l@{}}Adult \\ Census \citep{adult} \end{tabular} & 45175 & 14 & Gender & 1:3.0 & $>50K$ & Static \\
\hline
Compas \citep{compas} & 5278 & 9 &  Race &  1:1.1 & recidivism & Static \\
\hline
KDD \citep{kdd} & 299285 & 41 & Gender & 1:15.1 & $>50K$ & Static \\
\hline
Default \citep{default} & 30000 & 24 & Gender & 1:3.52 & default payment & Static \\
\hline
\begin{tabular}[c]{@{}l@{}}Law \\ School \citep{law} \end{tabular} & 18692 & 12 & Gender & 1:3.5 & pass bar & Static \\
\hline
NYPD \citep{nypd} & 311367 & 16 & Gender & 1:3.7 & felony & Stream \\
\hline
Loan \citep{loan} & 21443 & 38 & Gender & 1:1.26 & paid & Stream \\
\hline
\begin{tabular}[c]{@{}l@{}}Bank \\ Marketing \citep{bank} \end{tabular} & 41188 & 21 & \begin{tabular}[c]{@{}l@{}}Marital \\ Status\end{tabular} & 1:7.87 & subscription & Stream \\
\hline
\end{tabular}
\label{table1}
\end{center}
\end{table}

%As Naïve Bayes is actually based on the number of attributes 

\subsection{Evaluation metrics}
We evaluate the performance of the proposed model and its different modules using recall, balanced accuracy (B. Acc.), gmean, and cumulative statistical parity (Disc. score). 

\begingroup
\setlength{\tabcolsep}{5pt} % Default value: 6pt
\renewcommand{\arraystretch}{1} % Default value: 1
\begin{table}[t!]
\begin{center}
\small
\caption{Fairness and predictive performance of different variants of MNB and the baselines for streaming datasets, best and second best are in bold and underline respectively}
\begin{tabular}{p{1.8cm}p{2cm}p{1.7cm}p{1.7cm}p{1.7cm}p{1.8cm}}
\hline
Dataset & Model & Recall (\%)  & B.Acc. (\%) & Gmean (\%) & Disc. Score(\%) \\ \hline

\multirow{4}{*}{NYPD} 
& CSMOTE & \underline{97.29} &	57.49	& 41.47	 & 4.00 
  \\ 
& OSBoost &  \textbf{98.79} &	49.89 &	9.85 &	-0.34  \\
& MS & 17.47  & 56.93 & 41.06 & 5.87 \\  
& FAHT &  0.37 &	50.12 &	6.11 &	0.07  \\
& FABBOO & 46.83  & 62.96 & 60.78 & \underline{0.03}  \\  
 & MNB & 75.48 & \underline{77.34} & \underline{77.32} & 6.90\\  
 & DAMNB & 73.56  & 76.41 & 76.35 & 0.129 \\  
 & DCAMNB & 85.29 & \textbf{79.53} & \textbf{79.32} & \textbf{0.006} \\ \hline
 
 \multirow{4}{*}{\begin{tabular}[c]{@{}l@{}}Bank \\ Marketing \end{tabular}} 
 & CSMOTE & \textbf{85.61} &	\textbf{82.91} &	\textbf{82.87} &	8.29
  \\ 
& OSBoost &  36.45 & 66.85	& 59.49 & 3.04  \\ 
 & MS & 33.19  & 64.63 & 56.47  & 8.08\\ 
 & FAHT &  36.85 &	66.85 &	59.68 &	2.57 \\
 & FABBOO & 56.00 & 75.46 & 72.91 & 1.10\\   
   & MNB & 70.69 & 75.77 & 75.60 & 5.83 \\   
   & DAMNB & \underline{76.91}  & 75.68 & 75.67  & \textbf{-0.380}\\  
   & DCAMNB & 76.07  & \underline{78.95} & \underline{78.90}  & \underline{-0.579} \\ \hline
   
  \multirow{4}{*}{\parbox{1cm}{Loan}}  
  & CSMOTE & 74.77 & 69.17 & 68.94 & 3.58
  \\ 
& OSBoost & 77.21 &	68.71 &	68.18 &	5.56  \\ 
  & MS & 67.6  & 66.83 & 66.83  & 52.13\\ 
  & FAHT &  67.52 &	66.55 &	66.54 &	\underline{-0.34}  \\
  & FABBOO & 74.93  & 67.88 & 67.51  & 0.90\\   
   & MNB & \textbf{85.43}  & \textbf{77.81} & \textbf{77.43} & 5.34\\   
   & DAMNB & 85.18  & 77.49 & 77.11 & -0.393 \\  
   & DCAMNB & \underline{85.28}  & \underline{77.53} & \underline{77.14}  & \textbf{-0.295}\\ \hline
\end{tabular}
\label{table2}
\end{center}
\end{table}
\endgroup

\begingroup
\setlength{\tabcolsep}{5pt} % Default value: 6pt
\renewcommand{\arraystretch}{0.89} % Default value: 1

\begin{table}
\small
\begin{center}
\caption{Fairness and predictive performance of different variants of MNB and the baselines for static datasets, best and second best results are in bold and underline respectively}
\begin{tabular}{p{1.8cm}p{2cm}p{1.7cm}p{1.7cm}p{1.7cm}p{1.8cm}}

\hline
Dataset & Model & Recall (\%)  & B.Acc. (\%) & Gmean (\%) & Disc. Score (\%)  \\ \hline
\multirow{4}{*}{\parbox{1.2cm}{\centering Adult Census}}
& CSMOTE & \underline{80.46}	& 77.97 &	\textbf{77.91}	& 32.37
  \\ 
& OSBoost & 54.68 &	74.34 &	71.69 &	18.38  \\ 
& MS & 51.55 & 72.62& 69.45 &23.05  \\
& FAHT & 51.17&	72.62	& 69.38	& 16.37\\
& FABBOO & 64.53 & 75.11 & 74.36 & \underline{0.22} \\  
   & MNB & \textbf{76.66}  & \textbf{79.36} & \underline{79.31} & 38.94 \\  
   & DAMNB & 62.23  & 74.22 & 73.25 & \underline{0.22}\\  
   & DCAMNB & \textbf{76.38}  & \underline{78.61} & 77.85 &\textbf{0.045}\\ \hline

\multirow{4}{*}{KDD}  
 & CSMOTE & 64.98 &	75.8 &	74.13 &	11.23
  \\ 
& OSBoost & 31.64 &	65.37 &	55.97 &	3.95  \\ 
& MS & 24.91  & 62.02 & 49.71 & 15.80  \\ 
& FAHT & 27.9 &	63.47 &	52.41 &	2.93\\
& FABBOO & 77.98  & 81.48 & 77.98 & 0.13 \\  
   & MNB & 76.81  & \underline{81.83} & \underline{81.68} & 14.14\\  
   & DAMNB & \underline{82.92}  & 80.35 & 80.31 & \textbf{-0.069} \\  
   & DCAMNB & \textbf{86.81}  & \textbf{82.76} & \textbf{82.66} & \underline{-0.071} \\ \hline

 \multirow{4}{*}{\parbox{1cm}{Compas}} 
 & CSMOTE & 65.23 &	\textbf{65.35} &	\textbf{65.32} &	19.90
  \\ 
& OSBoost & 59.52 &	\underline{65.51} & \underline{65.23} &	27.46  \\ 
 & MS & 58.03   & 64.54 & 64.12 & 46.58  \\ 
 & FAHT & 60.39	& 64.57	& 64.37	& 22.05\\
 & FABBOO &  62.87 & 64.02 & 64.01 & 1.09  \\  
 & MNB & \underline{66.19}   & 64.76 & 64.74 & 43.83\\  
   & DAMNB & 65.25   & 63.03 & 62.94 & \textbf{-0.041}\\ 
   & DCAMNB & \textbf{68.98}  & 63.71 & 62.98 & \underline{0.254}\\ \hline

\multirow{4}{*}{Default} 
& CSMOTE & \textbf{80.58} &	59.1 &	55.03 & 2.46
  \\ 
& OSBoost & 31.16 & 63.22 & 54.46 & 2.11  \\ 
& MS & 30.54 & 63.08 & 54.01 & 11.15 \\ 
& FAHT & 30.61 &	62.99 &	53.99 &	1.71\\
& FABBOO & 41.4 & 65.94 & 61.17 & 0.98  \\  
   & MNB & 51.53  & 68.44 & 66.32 & 4.57\\  
   & DAMNB & 55.21  & \underline{68.78} &\underline{67.43} & \underline{0.09} \\  
   & DCAMNB & \underline{59.94}   & \textbf{69.01} & \textbf{68.41} & \textbf{0.023} \\ \hline

\multirow{4}{*}{\parbox{1 cm}{Law School}}  
& CSMOTE & \underline{77.88} &	\underline{76.67} &	\underline{76.66} &	2.16
  \\ 
& OSBoost & 16.51 &	57.53 &	40.31 &	1.65
  \\ 
& MS & 15.88   & 57.01 & 39.15 & 3.66 \\  
& FAHT & 9.62 &	54.33 &	30.5 &	0.87 \\
& FABBOO & 38.49  & 66.34 & 60.21 &0.31 \\   
   & MNB & 43.22 & 65.55 & 61.63  & 1.53\\  
   & DAMNB & 50.43  & 67.29 & 65.14 & \underline{0.042} \\   
   & DCAMNB & \textbf{80.38}  & \textbf{77.20} & \textbf{77.14} & \textbf{-0.017} \\ \hline
\end{tabular}
\label{table3}
\end{center}
\vspace{1cm}
\end{table}

\endgroup

\subsection{Experiments}
The proposed models are trained and tested following the prequential evaluation strategy, i.e., test first, then train. We tune the hyper-parameters $\alpha$, $\epsilon$, and $\lambda$ by grid search. To obtain the best results for all datasets, we choose values of $0.9$, $0.000001$, and $0.001$ for $\alpha$, $\epsilon$, and $\lambda$ respectively. As mentioned earlier, the non-streaming datasets lack temporal features; therefore, we use ten random shuffles of each static dataset and present the average of their results. 

\subsection{Results}
The measures of fairness and predictive performance obtained for a set of streaming datasets with different variants of the proposed MNB are presented in Table \ref{table2}. Similarly, the evaluation measures obtained on the average of $10$ random shuffles of each static dataset passed through the model are shown in Table \ref{table3}.  

\subsection{Impact of varying $\lambda$}
\begin{figure}[t] 
  \begin{subfigure}[b]{0.5\linewidth}
    \centering
    \includegraphics[width=\linewidth]{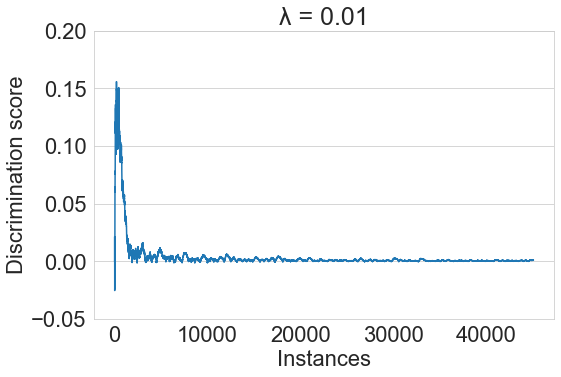} 
    \caption{}
    \label{fig4:a} 
  \end{subfigure}%% 
  \begin{subfigure}[b]{0.5\linewidth}
    \centering
    \includegraphics[width=\linewidth]{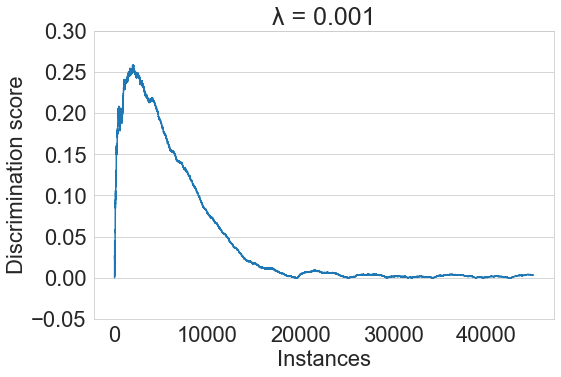} 
    \caption{}
    \label{fig4:b} 
  \end{subfigure} 
  \begin{subfigure}[b]{0.5\linewidth}
    \centering
    \includegraphics[width=\linewidth]{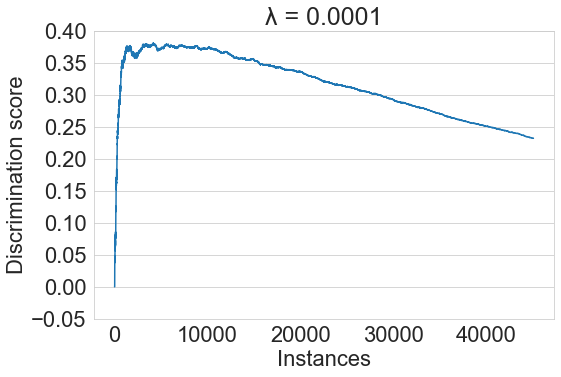} 
    \caption{}
    \label{fig4:c} 
  \end{subfigure}%%
  \begin{subfigure}[b]{0.5\linewidth}
    \centering
    \includegraphics[width=\linewidth]{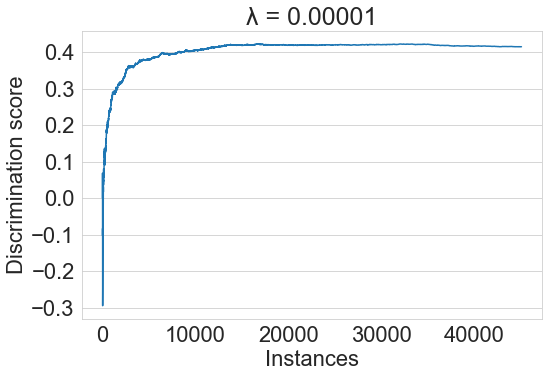}
    \caption{}
    \label{fig4:d} 
  \end{subfigure} 
  \caption{Impact of varying $\lambda$ on Disc. score (statistical parity) for Adult dataset}
  \label{fig4} 
\end{figure}

The most important hyper-parameter in reducing discrimination is $\lambda$ from the Algorithm \ref{alg:2}. We examined the effect of changing $\lambda$ on the ability of our proposed model to reduce discrimination, as shown in Figure \ref{fig4}. We use Adult Census dataset as a reference for this analysis. As can be seen in Figure \ref{fig4:a}, when the value of $\lambda$ is $0.01$, the discrimination value immediately drops to zero, indicating that this value is too large. With this value of $\lambda$, we achieve a balanced accuracy of $75.13\%$. If we decrease $\lambda$ to a value of $0.001$, the discrimination score decreases to a smaller and stable value after about $20,000$ instances, as shown in Figure \ref{fig4:b}. The balanced accuracy is also not much affected with a value of $78.61\%$. If we further decrease the value of $\lambda$ to $0.0001$, the discrimination score does not reach a stable value until the end of the stream, although it decreases as shown in Figure \ref{fig4:c}. This value of $\lambda$ leads to a balanced accuracy of $79.93\%$. As shown in Figure \ref{fig4:d}, if we leave $\lambda$ at $0.00001$, the discrimination score does not decrease throughout the data stream, and the achieved balanced accuracy is $80.37\%$. Therefore, we chose the value $0.001$ for $\lambda$, which provides a good trade-off between the balanced accuracy and the attenuation of the discrimination score.

\section{Discussion} \label{Cognizance behind the results}
We evaluate the effectiveness of our proposed model using both static and streaming datasets. From Tables \ref{table2} and \ref{table3} we can observe that we always achieve the best discrimination score as compared to all the baselines. Furthermore, we can observe that our predictive performance evaluation metrics experience a dip when we apply the discrimination mitigation strategy; this is the repercussion of this strategy. For instance, in case of the Adult Census dataset, the balanced accuracy drops from $79.36\%$ to $74.22\%$. However, we have shown that the balanced accuracy increases from $74.22\%$ to $78.61\%$ when we apply the class imbalance handling approach along with the discrimination mitigation method. Thus, our complete model "DCAMNB" demonstrates its multi-objective optimization behavior. Compared to state-of-the-art methods, the proposed methodology achieves the best balanced accuracy for Adult census, KDD, Default, Law School, NYPD, and Loan datasets. For Bank Marketing and Compas datasets the best balanced accuracy is achieved by CSMOTE which is not designed to handle discrimination but the proposed methodology DCAMNB closely follows  with just $3.96\%$ and $1.64\%$ decrease  in balanced accuracy. However, the difference between the discrimination score achieved by CSMOTE and proposed model is substantial; for Bank Marketing and Compas datasets CSMOTE achieves a discrimination score of $8.29\%$ $19.9\%$  whereas DCAMNB achieves  $-0.579\%$ and $0.254\%$ respectively. For the Compas dataset, FAHT and FABBOO achieved higher balanced accuracy, $64.54\%$ and $64.02\%$, respectively, than the proposed DCAMNB, i.e., $63.71\%$. However, the discrimination score achieved by DCAMNB ($0.254\%$) is substantially less as compared to those achieved by FAHT ($22.05\%$) and FABBOO ($1.09\%$).
However, we can observe that DCAMNB achieved an increase in balanced accuracy of $2.03\%$ for Adult Census dataset, $1.28\%$ for the KDD dataset, $16.57\%$ for the NYPD dataset, and $14.06\%$ for the Loan dataset. Similarly, we observe an increase of $1.28\%$, $4.68\%$, $1.52\%$, $18.54\%$, and $16.92\%$, in Gmean for the Adult Census, KDD, Default, NYPD, and Loan datasets, respectively. 
FABBOO's research question lies in close proximity with that of ours. It has the capability of reducing discrimination score to a suitable value while maintaining the balanced accuracy but FABBOO is not able to handle negative discrimination. Moreover, the predictive performance and discrimination scores achieved by the proposed methodology are much better than those of FABBOO. Compared to FABBOO's balanced accuracy, our proposed method provides an increase of $3.5\%$ for Adult Census dataset, $1.28\%$ for the KDD dataset, $3.07\%$ for Default dataset, $10.86\%$ for Law  School dataset, $16.57\%$ for the NYPD dataset, $3.49\%$ for Bank Marketing dataset, and $9.65\%$ for Loan dataset. Similarly, we observe an increase of $3.49\%$, $4.68\%$, $7.24\%$, $16.93\%$, $18.54\%$, $5.99\%$, $9.63\%$ in Gmean for the Adult Census, KDD, Default, Law School, NYPD, Bank Marketing and Loan datasets respectively as compared to that reported by FABBOO for these datasets.

\begin{figure}
  \centering
  \includegraphics[scale = 0.6]{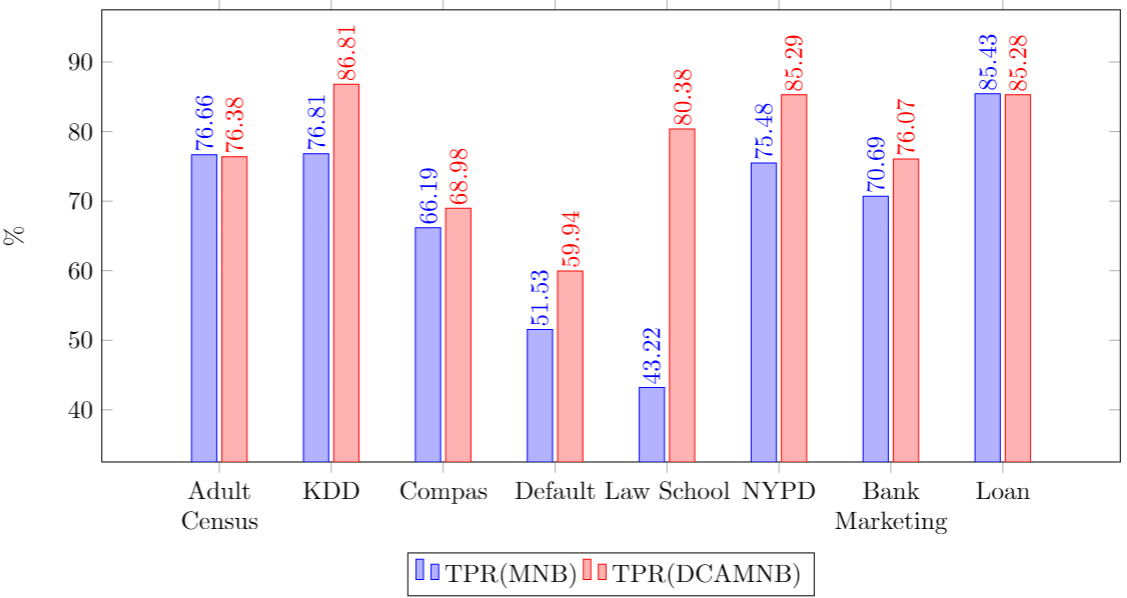}
  \caption{Comparison of TPR achieved by MNB and DCAMNB for Adult Census, KDD, Compas, Default, Law School, NYPD, Bank Marketing, and Loan datasets}
  \label{fig:2}
  \vspace{5mm}
  \includegraphics[scale = 0.6]{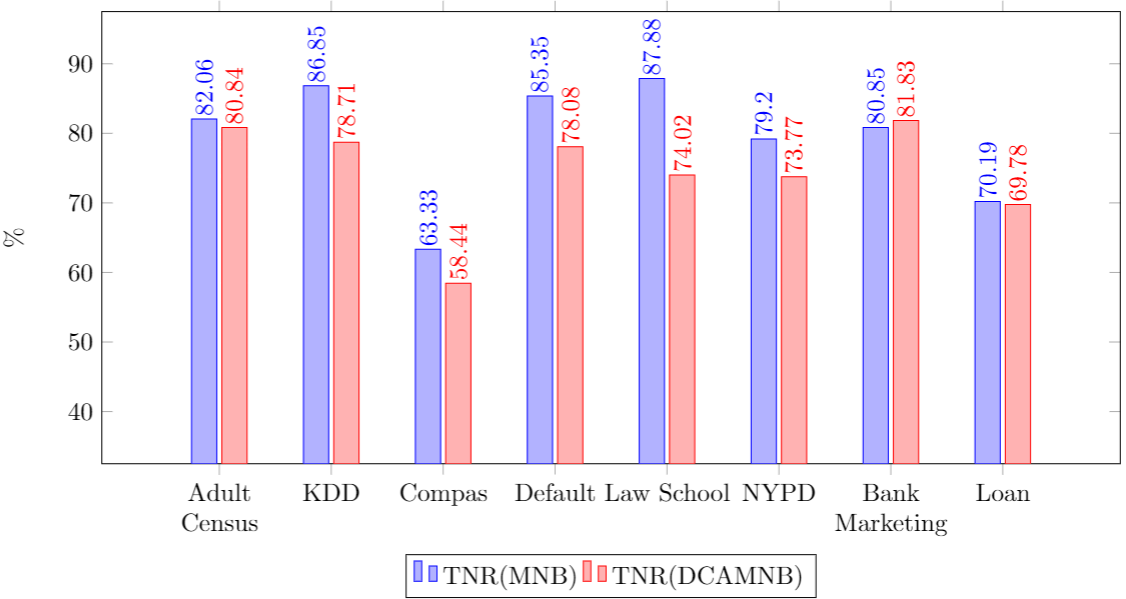}
  \caption{Comparison of TNR achieved by MNB and DCAMNB for Adult Census, KDD, Compas, Default, Law School, NYPD, Bank Marketing, and Loan datasets}
  \label{fig:3}
\end{figure}

Our instance weighting strategy has a pivotal positive effect on the true positive rate (TPR) while slightly compromising the true negative rate (TNR). For all imbalanced datasets, the TPR increases when we employ the instance weighting strategy together with the discrimination mitigation strategy. More specifically, we can observe that we achieve a $10$\% increase in TPR for the KDD Census dataset, $8.41$\% for the Default dataset, $37.16$\% for the Law School dataset, $9.81$\% for the NYPD dataset, and $5.38$\% for the Bank Marketing dataset, as shown in Figure \ref{fig:1}. However, the decrease in TNR for all datasets is much smaller than the increase in TPR. For the KDD dataset, we can observe a decrease in TNR of $0.923$\%, $7.27$\% for the Default dataset, $13.86$\% for Law School dataset, $5.43$\% for the NYPD dataset as depicted in Figure \ref{fig:2}. Moreover, for some datasets, we can even observe an increase in TNR, e.g., we can see an increase of $0.98$\% in TNR for the Bank Marketing dataset. The discrimination mitigation strategy severely impacted the model's predictive performance for the Adult Census dataset. The TPR dropped from $76.66$\% to $62.23$\%, and the balanced accuracy decreased from $79.36$\% to $74.22$\%. However, DCAMNB tried to compensate for the performance drop by boosting the TPR from $62.23$\% to $76.38$\% and increasing the balanced accuracy from $74.22$\% to $78.61$\%. Our method achieves parity between sensitive and non-sensitive groups even for balanced datasets, i.e., Compas and Loan.

\section{Conclusion}
The central prerequisite of a just and sustainable world is to ensure gender equality and realize the human rights, nobility, and competence of diverse groups of society. Therefore, we propose a discrimination-aware and class imbalance-aware online learning framework to achieve parity between favored and prejudiced groups of subjects.  

We present a novel adaptation of Naïve Bayes for mining data streams with embedded discrimination and class imbalance. To deal with class imbalance, we propose a unique weighting scheme which assigns a specific weight to each training object. We also propose a concept drift detection and handling module. Our approach mitigates both discrimination and reverse discrimination by online modifying the data distribution based on cumulative fairness notion. 
\\We have demonstrated the effectiveness of our methodology by conducting experiments on a range of static and streaming datasets. Our approach outperforms existing state-of-the-art methods in terms of both balanced accuracy and discrimination score. We have shown that our approach effectively learns both majority and minority classes and achieves a low discrimination score while maintaining high predictive performance. In contrast to state-of-the-art approaches, our method achieves low discrimination scores with substantially increased TPRs, while slightly compromising TNRs. 

In future work, we will thoroughly investigate the forgetting phenomena of the class imbalance handling module to make it adaptable to the nature of concept drift in the data. We also plan to analyse the theoretical aspects of our approach.
\\
\noindent\textbf{Ethics Statement}
Eliminating gender specific discrimination and eradicating exploitation of marginalised groups of society plays a decisive role in forging a path to a sustainable world. Our approach aims to make fair and high-quality predictions in accordance with the levels of moral equivalence; therefore, this methodology has the potential to address the issues of "gender equality" and "reducing inequalities". We use a range of datasets with "gender", "race", and "marital status" as the sensitive attributes to show the capability of the proposed model to address United Nation's Sustainable Development Goal 5 (gender equality) and 10 (reducing inequality)\footnote{\url {https://sdgs.un.org/goals}}.
\\
\noindent\textbf{Acknowledgements}
\noindent The authors acknowledge the support of the technical staff and faculty of the L3S Research Center at Leibniz University Hannover and the funding agency, the Lower Saxony Ministry of Science and Culture (Niedersächsische Ministerium für Wissenschaft und Kultur).
\bibliography{mybibfile}

\end{document}